\documentclass[10pt,twoside]{article}

\usepackage{times}
\usepackage[utf8]{inputenc}
\usepackage[T1]{fontenc}
\usepackage{graphicx}

\usepackage{siunitx}
\usepackage{taln2022}
\usepackage[french]{babel} 

\usepackage{subfig}
\title{Preuve de concept d'un bot vocal dialoguant en wolof}

\author{Elodie Gauthier\up{1}, Papa-Séga Wade\up{1}, Thierry Moudenc\up{1}, Patrice Collen\up{1}, Emilie De Neef\up{1}\quad Oumar Ba\up{2}, Ndeye Khoyane Cama\up{2}, Cheikh Ahmadou Bamba Kebe\up{2}, Ndeye Aissatou Gningue\up{2}\quad Thomas Mendo'o Aristide\up{3}\\
  {\small
    (1) Orange Innovation, France \\ 
    (2) Orange Sénégal, Dakar, Sénégal \\ 
    (3) ADNCorp, Dakar, Sénégal \\ 
    \texttt{
      \{elodie.gauthier, thierry.moudenc\}@orange.com \\ 
}}}

\begin{document}
\maketitle

\resume{
  Cet article présente la preuve de concept du premier assistant vocal automatique en wolof, première langue véhiculaire parlée au Sénégal. Ce bot vocal est le résultat d'un projet de recherche collaboratif entre Orange Innovation en France, Orange Sénégal (alias Sonatel) et ADNCorp, une petite société informatique basée à Dakar, au Sénégal. Le but du bot vocal est de fournir des informations aux clients d'Orange sur le programme de fidélité Sargal d'Orange Sénégal en utilisant le moyen le plus naturel de communiquer : la parole. Le bot vocal reçoit la demande orale du client, qui est traitée par un moteur de compréhension de la parole, et répond avec des messages audio préenregistrés. Les premiers résultats de cette preuve de concept sont encourageants : nous avons obtenu un WER de 22~\% pour la tâche de reconnaissance vocale et une F-mesure de 78~\% pour la tâche de compréhension.
}

\abstract{Proof-of-Concept of a Voicebot Speaking Wolof.}{
  This paper presents the proof-of-concept of the first automatic voice assistant ever built in Wolof language, the main vehicular language spoken in Senegal. This voicebot is the result of a collaborative research project between Orange Innovation in France, Orange Senegal (aka Sonatel) and ADNCorp, a small IT company based in Dakar, Senegal. The purpose of the voicebot is to provide information to Orange customers about the Sargal loyalty program of Orange Senegal by using the most natural mean to communicate: speech. The voicebot receives in input the customer's oral request that is then processed by a SLU system to reply to the customer's request using audio recordings. The first results of this proof-of-concept are encouraging as we achieved 22\% of WER for the ASR task and 78\% of F1-score on the NLU task. 
}

\motsClefs
  {bot vocal ; assistant virtuel ; wolof ; preuve de concept ; langues peu dotées ; reconnaissance automatique de la parole ; compréhension du langage naturel}
  {voicebot; virtual assistant; Wolof; proof-of-concept; low-resourced languages; automatic speech recognition; natural language understanding}

\section{Introduction}
Les chatbots sont aujourd'hui largement utilisés sur le Web et les assistants vocaux qui permettent de demander des informations sont de plus en plus présents au sein des foyers ou dans les voitures, au travers d'enceintes connectées ou de téléphones portables. Alors que les assistants vocaux sont très répandus dans les pays développés, ce n'est pas le cas dans les pays émergents. L'une des raisons est qu'ils nécessitent des technologies de traitement automatique des langues (TAL) qui ne sont pas assez matures pour les langues parlées dans ces pays. Les technologies de TAL ont besoin d'une quantité importante de données pour être performantes, mais les données dans les langues parlées dans les pays émergents sont très rares et ne sont pas faciles à collecter. \\
Au Sénégal, le français est la langue officielle. Néanmoins, comme dans de nombreux pays africains, la langue officielle du pays n'est parlée que par une minorité de la population (moins de 25~\% de la population sénégalaise). Cela pose de réels problèmes en termes d'inclusion et d'accès à l'information numérique. La majorité de la population sénégalaise parle et comprend le wolof (90~\% de la population, dont 40~\% de locuteurs natifs)\footnote{\url{http://www.axl.cefan.ulaval.ca/afrique/senegal.htm} (dernier accès: 13 mai 2022)}. 
Même si le wolof est bien documenté et décrit dans les études de linguistique \citep{senghor1947article,rialland2001intonational,mclaughlin2004there,cisse2006problemes,robert2011wolof,guerin2016constructions,voisin2021possession}, la langue souffre encore d'un manque de données numériques qui pourraient pourtant bénéficier au domaine du TAL. Le wolof est ainsi considéré comme une langue peu dotée. \\ 
Dans cet article, nous présentons le fruit d'une collaboration entre Orange Innovation, Orange Sénégal (alias Sonatel) et ADNCorp, une petite entreprise informatique basée à Dakar, au Sénégal.
Orange Sénégal est en train de développer un chatbot mais il n'interagit pour l'instant que par modalité textuelle et en français. Cela signifie que les clients sénégalais doivent être capables de lire et d'écrire en français. Or, plus de 50~\% des adultes sénégalais sont analphabètes \cite{unesco2017literacyrate}. Comme la population sénégalaise utilise le wolof dans ses interactions quotidiennes, le besoin d'un assistant automatique capable de dialoguer en wolof est de première importance. Nous avons donc choisi de développer le premier assistant vocal automatique jamais construit en langue wolof et ainsi permettre aux clients d'accéder à l'information par le biais de la parole. \\
\textbf{Contribution du papier.}
Cet article présente notre méthodologie pour concevoir et développer le premier bot vocal en wolof. Nous détaillons chacun des modules utilisés dans le système et les données que nous avons collectées à cet effet. Nous présentons les premiers résultats objectifs que nous avons obtenus sur des tâches de reconnaissance automatique de la parole (RAP) et de compréhension du langage naturel (NLU).\\
\textbf{Plan du papier.}
L'article est structuré de la façon suivante : tout d'abord, la section \ref{challenges} énumère certains des défis auxquels nous devons faire face lorsque nous traitons des langues d'Afrique subsaharienne (ASS), qui sont peu dotées. Ensuite, la section \ref{voicebot} présente le bot vocal en wolof que nous avons développé, en donnant quelques explications sur la gestion du dialogue. Puis, dans la section \ref{bot-units}, nous détaillons les modules qui composent l'assistant virtuel, de la RAP jusqu'à la réponse vocale. La section \ref{eval} présente les premiers résultats objectifs obtenus sur les tâches de RAP et de NLU qui constitue le moteur de compréhension du langage parlé (SLU) implémenté dans le bot vocal. Enfin, dans la section \ref{feedback} nous fournissons notre retour sur cette expérience tandis que la section \ref{conclusion} conclut ce travail et propose quelques perspectives sur nos activités de recherche.

\section{Principaux défis à relever lors du traitement des langues d'Afrique subsaharienne} \label{challenges}
Quand on travaille avec les langues d'ASS, de réels défis sont à considérer :
\begin{itemize}
	\item Les ressources numériques (orales ou écrites) sont très peu disponibles. 
	\item Les données qualitatives sont chronophages à collecter ou à produire. La formation des personnes pour transcrire des enregistrements est un processus très long et la tâche est loin d'être triviale, exigeant une grande rigueur. 
	\item Les langues d'Afrique subsaharienne sont à tradition orale : il n'existe généralement pas de règles pour écrire les mots (bien qu'il existe une orthographe officielle pour le wolof, elle est très peu connue et utilisée par les populations). 
	\item Les données de parole ne sont pas si coûteuses à collecter, mais le temps de transcriptions par l'homme est d'environ 8 fois supérieur.
	\item Les transcriptions de parole sont toujours importantes pour la RAP. Même si les travaux à l'état de l'art montrent des modèles en rupture qui n'ont pas (ou très peu) besoin de transcriptions d'enregistrements de parole pendant l'apprentissage \citep{carmantini2019LFMMI, synnaeve2020end, baevski2021unsupervised}, ces méthodes n'en sont qu'à leurs débuts. L'enthousiasme actuel pour les approches de bout-en-bout est séduisant et ces approches constitueront probablement la prochaine grande avancée dans le domaine puisque certains travaux ont déjà révélé le fort potentiel qu'elles pourraient apporter dans un contexte peu doté \citep{mohamud2021self,abate2021asr}. 
\end{itemize}
\vspace{-2mm}
\section{Le bot vocal en wolof} \label{voicebot}
\begin{figure}[h!]
    \centering
    \includegraphics[scale=0.08]{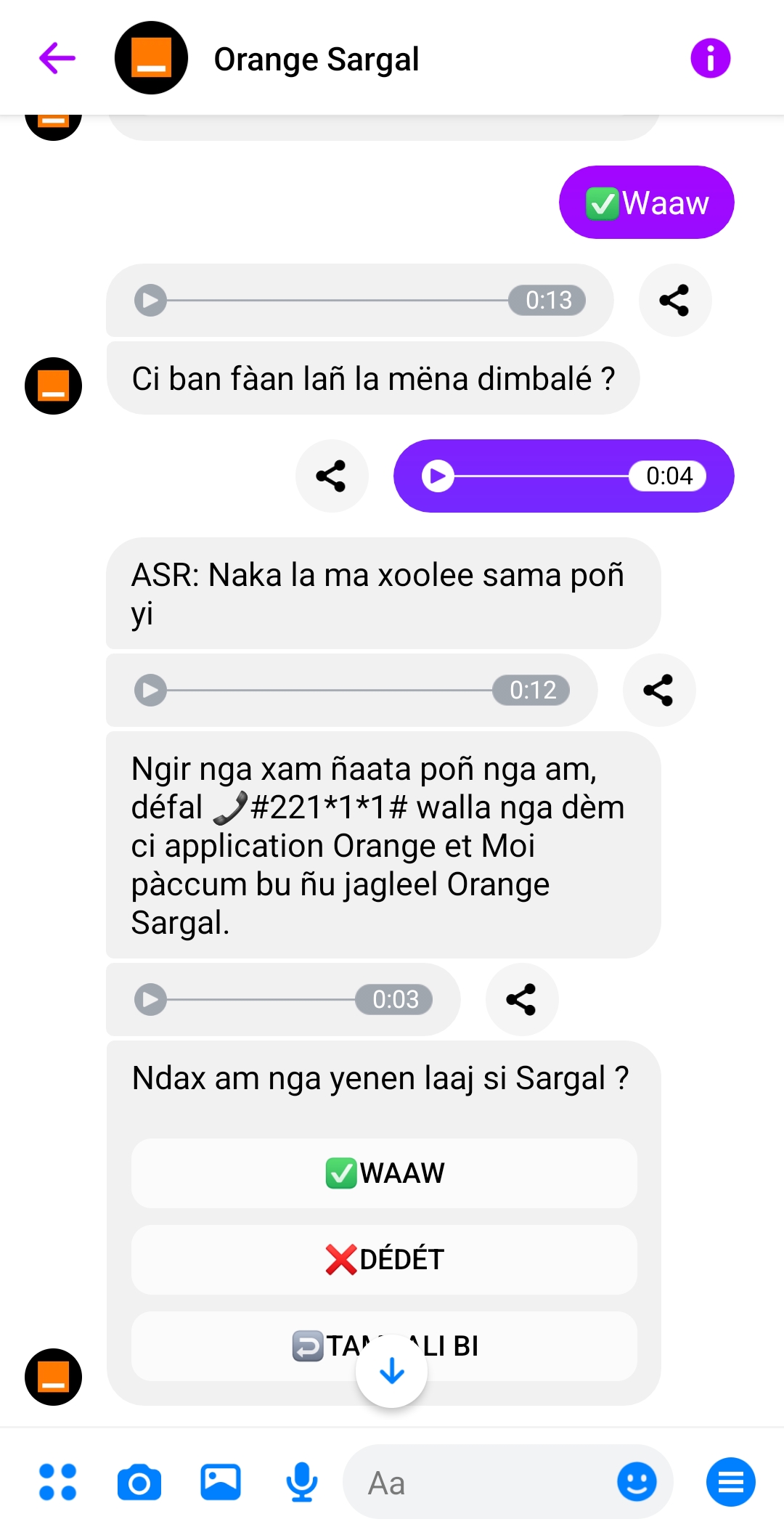}
    \caption{Capture d'écran d'un utilisateur interagissant avec le bot vocal, en wolof.\\ \small{Traduction du dialogue : l'\textbf{utilisateur} dit "Waaw" \textit{(Oui)} à une précédente interaction lors de laquelle le bot vocal a demandé à l'utilisateur s'il avait d'autres questions. Ensuite, le \textbf{bot vocal} répond : "Ci ban fàan lañ la mëna dimbalé ?" \textit{(Comment puis-je vous aider ?)}; l'\textbf{utilisateur} répond : "Naka la ma xoole sama poñ yi" \textit{(Comment puis-je consulter mes points ?)}. Le \textbf{bot vocal} répond : "Ngir nga xam ñaata poñ nga am, défal \#221*1*1\# walla nga dèm ci application Orange et Moi pàccum bu ñu jagleel Orange Sargal." \textit{(Pour consulter vos points, appelez le \#221*1*1\# ou allez sur l'application Orange et Moi, à la section Orange Sargal.)}. Le \textbf{bot vocal} ajoute: "Ndax nga yenen laaj si Sargal?" \textit{(Avez-vous d'autres questions sur Sargal ?)}. Cette question est suivie de trois boutons cliquables "WAAW" \textit{(Oui)}, "DÉDÉT" \textit{(Non)}, "TAMBALI BI" \textit{(Accueil)}.}}
    \label{fig:voicebot-screenshot2}
\end{figure}

La preuve de concept (PoC) du bot vocal cible les clients d'Orange Sénégal qui souhaitent accéder à des informations par la parole, en wolof, sur le programme de fidélité Sargal. \\ 
À ce jour, le voicebot est disponible sur Facebook Messenger et sur une interface web interne utilisée à des fins de développement et de tests. 
Le bot vocal est présenté dans une vidéo accessible sur la plateforme YouTube\footnote{\url{https://youtu.be/Vx3D3_8IOws/}}. \\
La figure \ref{fig:voicebot-screenshot2} montre une interaction entre un utilisateur et le bot vocal, sur l'application Facebook Messenger.
Une conversation entre un utilisateur et le bot vocal se déroule de la façon suivante : \\
\textbf{Initialisation de la conversation.} Avant toute chose, l'application Messenger envoie un message écrit à l'utilisateur pour lui indiquer que la session commence. 
Ensuite, le bot vocal initie la conversation en envoyant des salutations personnelles à l'utilisateur. \\ 
\textbf{Tour de parole de l'utilisateur.} Lors du tour de parole de l'utilisateur, celui-ci peut formuler sa question au bot vocal en utilisant le bouton microphone proposé par l'interface. 
À des fins de démonstration, nous affichons également la sortie du système de RAP à l'écran (elle apparaît dans la bulle qui commence par l'étiquette "ASR :", juste en dessous du widget de lecture violet encapsulant la demande orale de l'utilisateur, sur la figure \ref{fig:voicebot-screenshot2}).\\
\textbf{Tour de parole du bot vocal.} Lors du tour de parole du bot vocal, un message préenregistré répondant à la question de l'utilisateur est retourné\footnote{Nous avons choisi d'envoyer des préenregistrements car aucune information dynamique n'est nécessaire d'être apportée dans les réponses de ce cas d'usage. Les messages vocaux ont été enregistrés par une wolofone, à Dakar.}. L'interface Messenger propose un widget de lecture pour écouter les réponses vocales de l'assistant virtuel. En complément du message vocal, la transcription de l'enregistrement est affichée au-dessous\footnote{La transcription est évidemment réservée aux personnes alphabétisées en wolof (même si ce n'est pas la première catégorie d'utilisateurs visée par le voicebot) mais elle est cohérente avec le fait que l'interface Messenger ne permet pas aux développeurs de supprimer la saisie de texte. Le bot vocal autorise ainsi les interactions écrites.}.
À titre informatif, nous avons décidé de scinder les longues transcriptions d'un message audio en courtes transcriptions, car les messages successifs mais compacts sont plus faciles à suivre pour l'utilisateur. \\
\textbf{Fin d'un échange.} À la fin de l'échange\footnote{Dès que le bot vocal a fourni une réponse à l'utilisateur}, le bot vocal propose deux choix à l'utilisateur : poser une autre question ou retourner à l'écran d'accueil. \\
%
\vspace{-2mm}
\section{Composants principaux du bot vocal} \label{bot-units}
Afin d'identifier le sens (l'intention) de la requête de l'utilisateur, une tâche de compréhension de la parole est nécessaire.
Pour résoudre cette tâche, nous avons adopté l'approche qui consiste à coupler un moteur de RAP (pour transcrire la requête orale de l'utilisateur) à un moteur de NLU (pour extraire le sens de la sortie textuelle).
Une fois l'intention reconnue, elle est traitée par la brique de gestion du dialogue, qui choisit la réponse à délivrer à l'utilisateur de façon cohérente.

     \subsection{Le moteur de reconnaissance automatique de la parole}
\begin{figure}[h]
    \centering
    \includegraphics[scale=0.18]{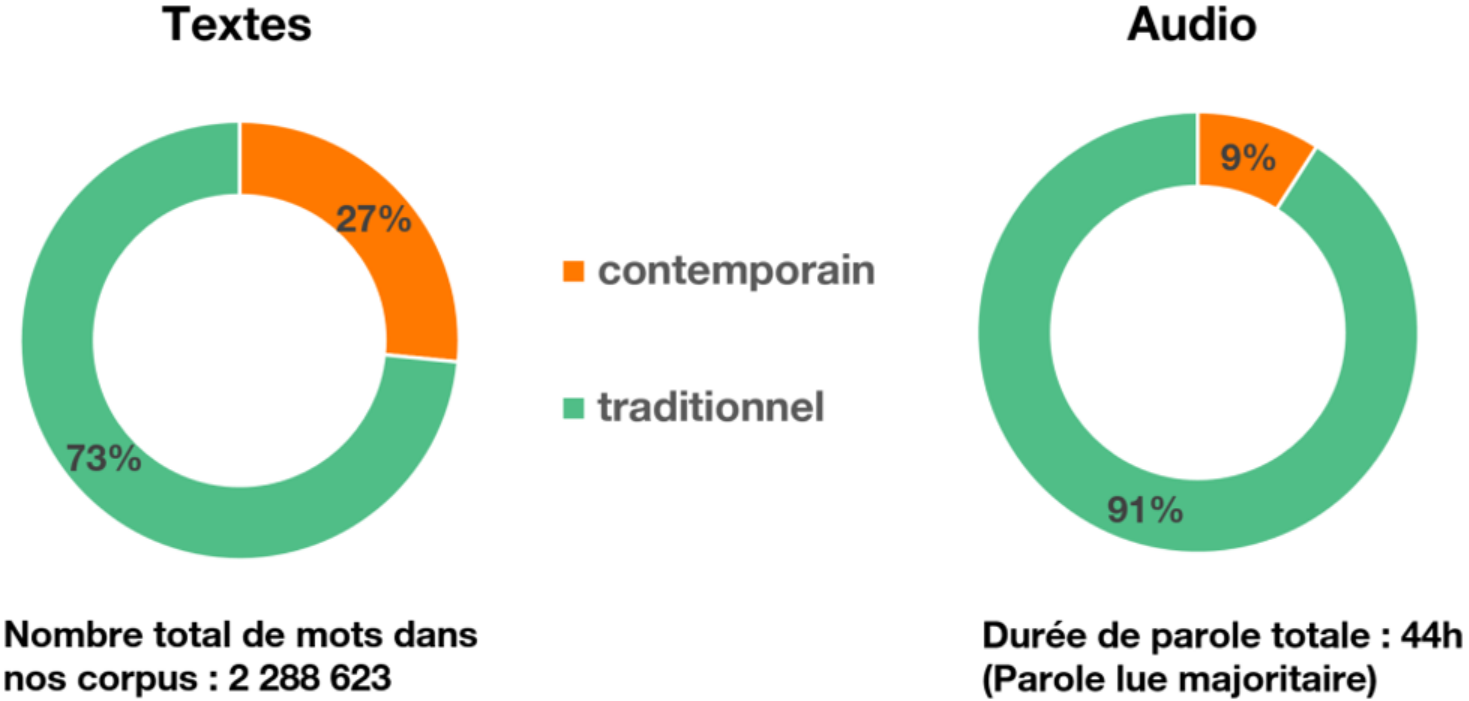}
    \caption{Répartition des jeux de données audio et textuelles utilisés pour entraîner la RAP du wolof.}
    \label{fig:ASR_data_PoC}
\end{figure}

    La tâche de reconnaissance automatique du wolof est accomplie par la technologie de reconnaissance vocale d'Orange qui repose sur certains outils fournis par Kaldi \cite{povey2011kaldi}.\\
        \textbf{Données.} Les jeux de données que nous avons utilisés pour entraîner le système de RAP contiennent principalement du wolof dit traditionnel (voir figure \ref{fig:ASR_data_PoC}). C'est en effet le genre de contenu que l'on retrouve majoritairement sur le Web\footnote{Nous \oe{}uvrons actuellement à pallier ce problème en collectant des documents contemporains, dans un wolof plus proche de celui parlé en réalité par la population sénégalaise, plus adapté à notre cas d'usage.}.
        Le lexique de prononciation contient 50\,271 entrées. L'alternance codique français-wolof étant fréquente au Sénégal \citep{thiam1994variation,nunez2017pratiques,sarr2017plurilinguisme,diagne2018terminologie}, environ 4k de mots français ont été inclus. Les transcriptions phonétiques sont encodées en SAMPA et ont été générées automatiquement à l'aide de notre système de conversion graphèmes-à-phonèmes développé en interne.
        2 millions de mots (issus principalement de contenus en wolof traditionnel) ont été utilisés pour estimer le modèle de langue. Nous avons utilisé des textes écrits avec l'orthographe officielle basée sur l'alphabet latin. Nous sommes conscients que le wolof peut également être écrit avec une déclinaison de l'alphabet arabe (appelée wolofal) mais l'absence de clavier standardisé limite la production de corpus (notamment pour la transcription d'enregistrements de parole). Pour cette raison, nous n'avons pas privilégié le wolofal pour le moment. 
        44 heures de parole, majoritairement propre et lue, sur laquelle nous avons appliqué des méthodes d'augmentation de données pour tripler la quantité de données d'apprentissage, ont été utilisées pour entraîner la version du modèle acoustique actuellement présentée dans cet article et implémentée dans le bot vocal. 
        Pour toutes les itérations de test du système de RAP, nous avons utilisé un jeu composé d'environ 30 minutes de discours propre et lu en wolof, où les structures lexicales et syntaxiques sont conformes à la façon dont les Sénégalais parlent aujourd'hui. \\
        \textbf{Configuration.} Le système de RAP développé pour transcrire les mots prononcés par le client utilise une architecture hybride classique. Le modèle acoustique possède environ 15 millions de paramètres et a été entraîné à l'aide d'un réseau de neurones à retard temporel (TDNN) avec 7 couches cachées. Chaque couche est composée de 625 neurones et la couche de sortie possède 5\,936 neurones. 
        Le modèle de langue estimé, d'une taille est d'environ 35Mb, est un modèle 5-grammes. 2 cartes graphiques (Nvidia Tesla P100) ont servi à l'apprentissage du modèle acoustique, qui a duré environ 18 heures.
    \subsection{Les moteurs de compréhension du message et de gestion du dialogue}
    La tâche de compréhension du message et la gestion du dialogue sont réalisées à l'aide du framework open source Rasa\footnote{\url{https://rasa.com/}}.\\ 
        \textbf{Données.} Nous devions utiliser des données réelles provenant des centres d'appels d'Orange Sénégal pour alimenter le PoC mais des questions d'ordre juridique sur la réglementation de la protection des données ont retardé nos expériences.
        En se basant sur un référentiel de 30 intentions préalablement identifiées par les ergonomes d'Orange au Sénégal et d'Orange Innovation en France, nous avons alors simulé un ensemble d'énoncés reflétant le plus possible une interaction réelle entre un client et un téléconseiller. 
        Nous avons ainsi constitué un jeu de données équilibré contenant plusieurs énoncés répartis en 9 classes d'intentions, grâce à des volontaires wolofones à Dakar.\\ 
        \textbf{Configuration du modèle de compréhension du message.} 
        Rasa met à disposition différents types de composants qui fonctionnent de façon séquentielle dans la chaîne de compréhension du message. \\
        Le modèle de compréhension est entraîné à partir de l'algorithme \textit{DIETClassifier} (Dual Intent and Entity Transformer, \citet{mantha2020diet}) qui exploite des expressions régulières, définies dans les données d'entraînement, pour extraire les entités et classer les intentions présentes dans chaque exemple de message. Le classifieur DIET que nous avons configuré prend en entrée des vecteurs de caractéristiques clairsemées (\textit{sparse features} en anglais) qui proviennent des composants \textit{LexicalSyntacticFeaturizer} et \textit{CountVectorsFeaturizer}\footnote{Pour une description détaillée de ces composants, se référer à la documentation de Rasa : \url{https://rasa.com/docs/rasa/components/}}. \\
        \textbf{Configuration du modèle de gestion du dialogue.} Le modèle de gestion du dialogue proposé par Rasa permet au bot de décider de l'action à exécuter au fil de la conversation au moyen de procédures préconfigurées. Ces procédures permettent de prédire la ou les prochaines actions et sont automatiquement sélectionnées à chaque tour de parole. Les algorithmes de ces procédures peuvent être basés sur de l'apprentissage automatique ou sur des règles. Une seule procédure à la fois, sélectionnée en fonction d'un score de confiance, peut déclencher l'action. Nous avons configuré 3 procédures dans le bot vocal parmi celles proposées par le framework (classées de la priorité la plus forte à la plus faible accordée par Rasa) : \textit{RulePolicy} (cette procédure utilise un algorithme à base de règles), \textit{AugmentedMemoizationPolicy} et \textit{TEDPolicy} \citep{vlasov2019dialogue} (ces deux dernières sont basées sur des algorithmes d'apprentissage automatique)\footnote{Pour une description détaillée de ces composants, se référer à la documentation de Rasa : \url{https://rasa.com/docs/rasa/policies/}}. \\
\vspace{-2mm}
    \subsection{Réponse vocale}
    Nous avons choisi de répondre au client par des messages audio préenregistrés. Les messages sont prononcés par une locutrice de wolof. Comme aucun paramètre dynamique n'est nécessaire pour produire les réponses, nous avons préféré enregistrer les différentes réponses, assurant ainsi une bonne expérience utilisateur.
    \vspace{-2mm}
\section{Premières évaluations\label{eval}}
    \subsection{Performance de la RAP}
    La brique de reconnaissance automatique de la parole a nécessité plusieurs itérations de développement avant d'atteindre une performance respectable. 
    La toute première version a été construite en utilisant les jeux de données du projet ALFFA \cite{gauthier2016lrec}.  
    Lors de la première itération, nous avons utilisé l'ensemble d'entraînement (16h d'enregistrements de parole nettoyés et transcrits), le lexique de prononciation d'environ 32k entrées transcrites phonétiquement en SAMPA et le modèle de langue estimé (à partir d'environ 600k mots) disponibles sur le dépôt GitHub du projet ALFFA\footnote{\url{https://github.com/getalp/ALFFA_PUBLIC/tree/master/ASR/WOLOF/}} pour apprendre les différents modèles de RAP. Avec cette configuration limitée aux données du projet ALFFA, nous avons obtenu un taux d'erreur mot (WER) de 70~\% sur un ensemble de test en wolof lu contemporain. Ce premier résultat représente notre base de référence.
    Nous avons ensuite produit une nouvelle version en augmentant le modèle de langue avec davantage de données textuelles collectées sur le Web à partir de contenus Wikipedia, de sites web sénégalais, de journaux en ligne ainsi que de transcriptions de contenus d'émissions de radio et de télévision qui ont été produites par deux linguistes wolofones au Sénégal. Cette configuration nous a permis d'atteindre un gain relatif sur le WER de 11~\%, sur le même ensemble de test utilisé pour la base de référence.
    Enfin, la dernière version en date\footnote{Actuellement implémentée dans le PoC présenté dans cet article.} a été développée en ajoutant du vocabulaire wolof et français dans le lexique de prononciation. Il dénombre désormais 50k entrées environ, dont près de 4k mots français. Avec cette configuration, nous avons amélioré les performances du modèle entraîné en obtenant un WER de 22~\% sur l'ensemble de test.\\ 
    Nous sommes continuellement en train d'améliorer les modèles de RAP. Nous sommes aujourd'hui en passe d'intégrer des transcriptions des enregistrements des centres d'appels d'Orange Sénégal\footnote{Ces données ont été préalablement anonymisées afin d'être conformes au Règlement général sur la protection des données.} pour améliorer le modèle de langue. Nous prévoyons à terme d'intégrer ces enregistrements à nos modèles acoustiques. 
    
    \subsection{Performance du NLU}
        Comme nous avons très peu d'échantillons par classe d'intentions dans notre jeu de données (avec une moyenne de vingt échantillons par classe), nous avons utilisé une validation croisée à 5 blocs pour tester nos modèles NLU. 
        Au total, nous disposons de 184 observations réparties en 9 classes.
        Le modèle NLU obtient de bons résultats sur les scénarios simulés. 5 intentions sur 9 obtiennent une F-mesure supérieure à 80~\%. Le score minimal obtenu parmi les 4 autres intentions est de 57~\%.
        Dans cette première itération, nous avons obtenu une moyenne pondérée de 79~\% pour le rappel, 79~\% pour la précision et 78~\% pour la F-mesure. 
        Ces résultats sont encourageants pour la prochaine étape qui consistera à expérimenter sur des données métier provenant des centres d'appels d'Orange Sénégal.
\vspace{-2mm}    
\section{Test en condition réelle et analyse préliminaire}
    Nous avons réalisé une campagne de tests du PoC en condition réelle, grâce à 40 salariés d'Orange Sénégal qui se sont portés volontaires pour réaliser le test (de type FUT -- \textit{Friendly Users Test} en anglais). Le but d'un FUT est notamment de collecter un maximum de données réelles pour augmenter les corpus et renforcer les modèles en vue de bêta-tests avec des utilisateurs externes à l'entreprise.
    Afin que les tests se déroulent le plus uniformément possible entre tous les testeurs, nous avons partagé des consignes d'interaction. 
    En fin de test, 15 testeurs ont répondu à un questionnaire de satisfaction. Ils ont attribué à la qualité de la compréhension une note moyenne de 3/5. \\
    \textbf{Analyse des sorties du système de SLU obtenues pendant le FUT.} 200 messages prononcés par les utilisateurs ont été recueillis pendant le FUT. Alors que 192 messages prononcés n'ont pas été correctement transcrits par le moteur de RAP, le moteur NLU a correctement identifié 30~\% des intentions présentes dans les messages (60 intentions correctement identifiées sur 200 requêtes), ce qui signifie que le moteur de NLU est capable de relativement bien généraliser les contextes d'intentions.
    Ces résultats préliminaires montrent que, dans un contexte d'utilisation réel, les performances sont beaucoup plus faibles que ceux des modèles évalués séparément au cours des expérimentations. Cependant, des analyses qualitatives supplémentaires doivent être menées notamment afin de spécifier les types d'erreurs du système de RAP. \\
    \textbf{Profil des testeurs.} La majorité du panel de testeurs connaissait le programme de fidélité Sargal (62\% des testeurs étaient même inscrits au programme) et utilisait Facebook Messenger de façon occasionnelle. Nous sommes conscients que les bêta-tests qui suivront devront s’orienter vers un panel plus représentatif de la population sénégalaise (moins à l’aise avec le numérique et dont l’interaction orale, en langue locale, a un impact significatif sur l’appétence de tels services) afin de tirer des enseignements substantiels et de s’assurer de l’acceptabilité d’une telle solution.
\vspace{-2mm}
\section{Retour sur expérience} \label{feedback}
Nous avons collaboré avec des entreprises sénégalaises pour augmenter nos ensembles de données de parole en wolof. Elles ont partagé avec nous des enregistrements d'émissions en wolof qui ont été transcrites par des linguistes. Les transcripteurs parlent et écrivent le wolof au quotidien. Au-delà du choix des transcripteurs, l'homogénéité des transcriptions est essentielle. Nous avons donc donné quelques règles d'annotation à suivre aux transcripteurs. La principale difficulté que nous avons rencontrée est la transcription des données qui est une tâche très difficile lorsqu'il s'agit de langues locales parlées plutôt qu'écrites, même pour les transcripteurs ayant une formation de haut-niveau en linguistique. Plusieurs cycles de révision ont dû être mis en place avant que les transcriptions ne soient propres.
\vspace{-2mm}
\section{Conclusion} \label{conclusion}
Dans cet article, nous présentons le premier assistant virtuel dialoguant en wolof par le biais de la parole. Au Sénégal, le taux d'alphabétisation des adultes est encore faible (moins de 50~\%) et le français n'est pratiqué quotidiennement que par un quart de la population. Construire un bot vocal capable d'interagir en wolof avec nos services est un enjeu majeur. 
Le bot vocal prend en entrée un signal de parole et retourne une réponse au client à travers des enregistrements audio. L'échange de messages écrits est également possible afin de ne pas dégrader l'expérience client, car l'interface Facebook Messenger ne propose pas de supprimer le module de saisie de texte. Une première série de tests ont été réalisé en situation d'usage réel et de nouveaux tests utilisateurs sont prévus à l'été 2022.
Nous prévoyons également de mettre à niveau le système de RAP en augmentant la quantité de données d'apprentissage et en améliorant leur qualité. Nous comptons améliorer le modèle acoustique avec des variétés accentuées de français africain, ajouter plus de mots français et de noms de marques d'Orange dans le lexique de prononciation mais aussi étendre le modèle de langue avec des expressions plus contemporaines qui reflètent davantage la façon de parler, les expériences et les cultures de la population sénégalaise. 
Enfin, nous aspirons à intégrer un composant de génération automatique de texte (NLG) couplé à un système de synthèse vocale (TTS) en wolof  pour générer des réponses à la volée et offrir un comportement dynamique à l'utilisateur. Le composant NLG permettra de générer une variété de réponses textuelles qui seront lues à haute voix par le système TTS au lieu des réponses préenregistrées actuellement proposées.
%

\bibliographystyle{taln2022}
\bibliography{biblio}

\end{document}